\documentclass[letterpaper, 10 pt, conference]{ieeeconf}  

\IEEEoverridecommandlockouts                              

\usepackage{graphics} 
\usepackage{epsfig} 
\usepackage{mathptmx} 
\usepackage{times} 
\usepackage{amsmath} 
\usepackage{amssymb}  
\usepackage{bbm}
\usepackage{cuted}
\usepackage{caption}

\usepackage[colorlinks,linkcolor=blue]{hyperref}

\title{\LARGE \bf
Playful DoggyBot: Learning Agile and Precise Quadrupedal Locomotion
}

\author{
Xin Duan$^{1}$$^{2}$ Ziwen Zhuang$^{1}$$^{3}$
Hang Zhao$^{1}$$^{3}$ Sören Schwertfeger$^{2}$ \thanks{
$^{1}$~Shanghai Qi Zhi Institute \quad $^{2}$~Key Laboratory of Intelligent Perception and Human-Machine Collaboration -- ShanghaiTech University. \quad $^{3}$~Tsinghua University}
}

\begin{document}

\maketitle
\thispagestyle{empty}
\pagestyle{empty}

\begin{strip}
\begin{minipage}{\textwidth}\centering
    \vspace{-1.9cm}
    \includegraphics[width=0.9\textwidth]{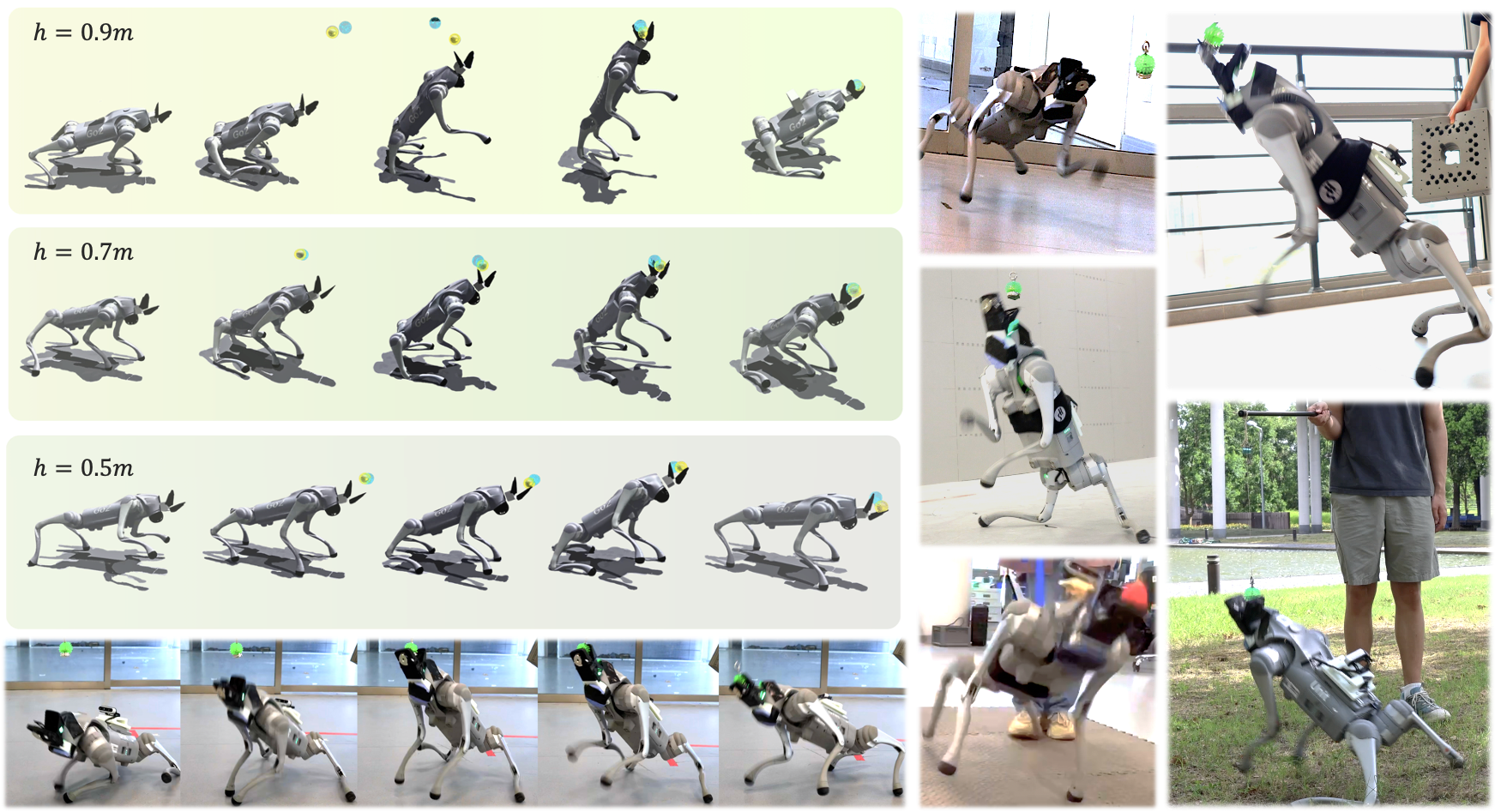}
    \captionof{figure}{\textbf{Playful DoggyBot.} We present a system to explore the agile and precise movements of quadrupedal robots. A robot dog mounted with a mouth-like gripper can finish the challenging task of leaping up to catch a small target object. Code and videos are available on the Project Webpage: \url{https://playful-doggybot.github.io/}.\vspace{-.2cm}}
    \label{fig:main image}
\end{minipage}
\end{strip}

\begin{abstract}

Quadrupedal animals can perform agile and playful tasks while interacting with real-world objects. For instance, a trained dog can track and catch a flying frisbee before it touches the ground, while a cat left alone at home may leap to grasp the door handle. Successfully grasping an object during high-dynamic locomotion requires highly precise perception and control. However, due to hardware limitations, agility and precision are usually a trade-off in robotics problems. In this work, we employ a perception-control decoupled system based on Reinforcement Learning (RL), aiming to explore the level of precision a quadrupedal robot can achieve while interacting with objects during high-dynamic locomotion.
Our experiments show that our quadrupedal robot, mounted with a passive gripper in front of the robot's chassis, can perform both tracking and catching tasks similar to a real trained dog. The robot can follow a mid-air ball moving at speeds of up to 3m/s and it can leap and successfully catch a small object hanging above it at a height of 1.05m in simulation and 0.8m in the real world.


\end{abstract}

\section{INTRODUCTION}
A common scenario in our daily lives features a trained dog
expertly chasing after a fast-moving frisbee and leaping up to catch it just before it hits the ground.
The ability of robots to perform agile movements and precise manipulation tasks in dynamic environments is a crucial aspect of their functionality.
Successfully catching the objects requires real-time visual perception and the ability to remember and understand the surrounding environment~\cite{patla1997understanding}. Visual perception and memory enable the dog to detect and track the trajectory of the target object, and then to 
accurately judge distance and timing for the leap, knowing the optimal positioning and torque required.
However, the ability of robots to act quickly and accurately with real-world objects remains a challenging task that requires exceptional agility and precision.
In this work, we equip a quadruped robot known for its agile movements with an end-effector resembling a dog's mouth and train it to track, jump up, and catch a mid-air ball smaller than a tennis ball (with a diameter of no more than 5 centimeters), aiming to explore its agility and precision. The robotic dog demonstrates continuous pursuit capability while the small ball is rapidly and randomly moving suspended from a string and operated by a human.

During high-speed locomotion of legged robots, sensor inaccuracies and latency as well as actuator execution errors become noticeable.
In our configuration, the egocentric vision system experiences both rapid horizontal motion and persistent vertical jolts from the legged robot's dynamic gait, which induces pronounced motion blur in captured images, thereby reducing target object detection accuracy.
At $1.5 m/s$ locomotion speed, even a small unexpected delay of $0.05 s$ will lead to $0.075 m$ positional error, which already exceeds the accuracy requirement in our task. 
Such conditions require that onboard system be able to adjust its behavior in real-time.
In contrast, manipulation tasks demand greater operational accuracy, but they are typically deployed in quasi-static environments where sensor-induced errors are minimal, not requiring real-time computation. Thus, these tasks can utilize complex, large models in the system. 
Recent studies tried to resolve this issue, but focused on simpler tasks. In~\cite{Huang2022CreatingAD}, the quadrupedal robot uses its legs to interact with a soccer ball that has a relatively large diameter of approximately $20 cm$, which does not require extremely high accuracy. Although~\cite{Forrai2023event} employs a tennis ball-sized object, the catching device remains relatively large, thereby allowing greater margin for error. \cite{wu2024helpful} disentangles the agile maneuvering from precise manipulation through two operation stage by moving to the target with agile locomotion and perform the manipulation in a relatively slow manner, so that these two technical requirement do not conflict with each other. Some tasks require both agility and precision \cite{zhuang2023robot,cheng2023parkour}, but the robot can resolve the trade-off with more action redundancy. For instance, when a quadrupedal robot is trained to climb a certain staircase, it can lower its precision requirements by just lifting its legs higher. The policy learns to act with more torques or take the leap slightly earlier than the exact acting point. 
In our work, we aim to solve the tracking and catching tasks without much room for action redundancy. 
Successfully catching demands the small ball to avoid being knocked away by the gripper or other body components of the robot and hit a rectangular catching region that is less than $25cm^2$. 
The robot cannot improve its hit rate by jumping higher or lower, which would knock the ball away by hitting the edges of the gripper. 

The main contributions are as follows:
\begin{itemize}
    \item We demonstrate a complete system for high-speed and precise quadrupedal manipulation, enabling a real robot to track small objects moving at speeds of up to 3m/s with sudden direction changes, and successfully catch targets at heights of up to 0.8m while in dynamic motion.
    
    \item We employ a comprehensive training framework that addresses the agility-precision trade-off through carefully designed reward functions, curriculum learning, and memory-equipped network architectures.
    
    \item We validate our simulation-trained policies in simulation and real-world experiments. We analyze key factors affecting catching performance, including the perception noise of the target position, the absolute height of the target and the number of different target heights during training, providing insights into the challenges of dynamic object manipulation for quadrupedal robots.

\end{itemize}


\section{RELATED WORK}
\subsection{Legged Agile Locomotion}
There is lots of work like the ETH STARLETH robot~\cite{7414416}, Boston Atlas robot~\cite{atlas} and the MIT Cheetah robot~\cite{nguyen2019optimized} showing impressive legged locomotion ability including walking on various terrain~\cite{5509176,pmlr-v78-antonova17a,DBLP:ZuckerRSCBAK11,Kolter2011TheSL,5509805,doi:10.1126/scirobotics.abk2822,9779429}, running~\cite{shen2022legged}, jumping~\cite{7414416}, stepping over obstacles~\cite{park2015online,nguyen2019optimized,nguyen2022continuous} and even smooth parkour skills~\cite{atlas}. However, these cases using model-based control techniques usually need a lot of engineering efforts for modeling the robots and the environment and suffer from scaling its ability to diverse environments and changes in dynamics. Recently, we have experienced an explosive development of learning-based control techniques that not only accomplish the basic ability we mentioned before~\cite{margolisyang2022rapid,ji2022concurrent,fu2021minimizing,kumar2021rma,nguyen2022continuous,margolis2021learning}, but also various fancy locomotion skills including climbing up and down stairs~\cite{lee2020learning,nahrendra2023dreamwaq,agarwal2022legged,yang2023neural,Loquercio2023visual,hoeller2022neural,Rudin2022advanced,rudin2022learning}, resetting to safe gaits~\cite{hwangbo2019learning}, jumping over gaps~\cite{Rudin2022advanced,zhuang2023robot,cheng2023parkour}, back-flipping~\cite{Li2022LearningAS}, standing up on rear legs~\cite{Laura2023learning,long2024hinf}, moving with damaged parts~\cite{Nagabandi2018LearningTA}, weaving poles~\cite{Caluwaerts2023BarkourBA} and also parkour skills~\cite{zhuang2023robot,cheng2023parkour}. There are also combinations of model based control and Deep Reinforcement Learning (DRL), leveraging the advantages of both to improve robustness and generalization~\cite{fabian2024dtc,margolis2021learning}.

\subsection{Quadrupedal Manipulation}
Locomotion focuses more on the robot's own movement capabilities, while manipulation focuses on the robot's ability to interact with objects in the real world.
The most direct way to enable quadrupedal robots to do manipulation tasks involves mounting an arm manipulator on it~\cite{arcari2023bayesian,ha2024umilegsmakingmanipulation,fu2022deep,portela2024learning,pan2024roboduet,ito2022efficient}. Since sometimes the four limbs for the quadrupedal robot are redundant for walking, there are also some works using legs as manipulators~\cite{cheng2023legs,he2024learning,Arm2024PedipulateEM,Ji2023dribblebot,Huang2022CreatingAD,ji2022hierarchical} or with a gripper mounted on the leg~\cite{he2024learning,lin2024locoman,Arm2024PedipulateEM}. Inspired by the morphology of dogs in nature, we mount a gripper on the head position to enable simple grasp actions.

Previous work has shown the interaction ability, including opening doors~\cite{cheng2023legs,he2024learning,fu2022deep,ito2022efficient}, pressing buttons~\cite{cheng2023legs,Arm2024PedipulateEM,he2024learning}, and picking and placing objects~\cite{Arm2024PedipulateEM,fu2022deep}.~\cite{Huang2022CreatingAD} works as a soccer goalkeeper to stop a ball flown over using a third-view camera.~\cite{Forrai2023event}  catches the small flying ball with a net bag using an event camera, requiring the robot to  move to the specified location before the key time point. In both of these works, the robot initially stays stationary, waiting until the estimator calculates the ball's landing point before it moves to the designated position at the critical time step. Similar to our approach~\cite{wu2024helpful} mounts a 1-DoF gripper to the front of a quadruped, serving as the end-effector. They leverage pre-trained vision-language models (VLMs) to develop a robot system for learning quadrupedal mobile manipulation. Following human commands, the robot performs tasks like climbing onto the bed and then picking up the desired object. 
In this work, we test the possibility of a robot performing extremely precise action while moving at a high speed with only its onboard sensors and computation.
Therefore, our robot dog is equipped with an end effector to run toward the target and capture it.

\section{METHOD}
\begin{figure*}[ht!]
\centering
\vspace{-0.1cm}
\includegraphics[width=0.9\linewidth]{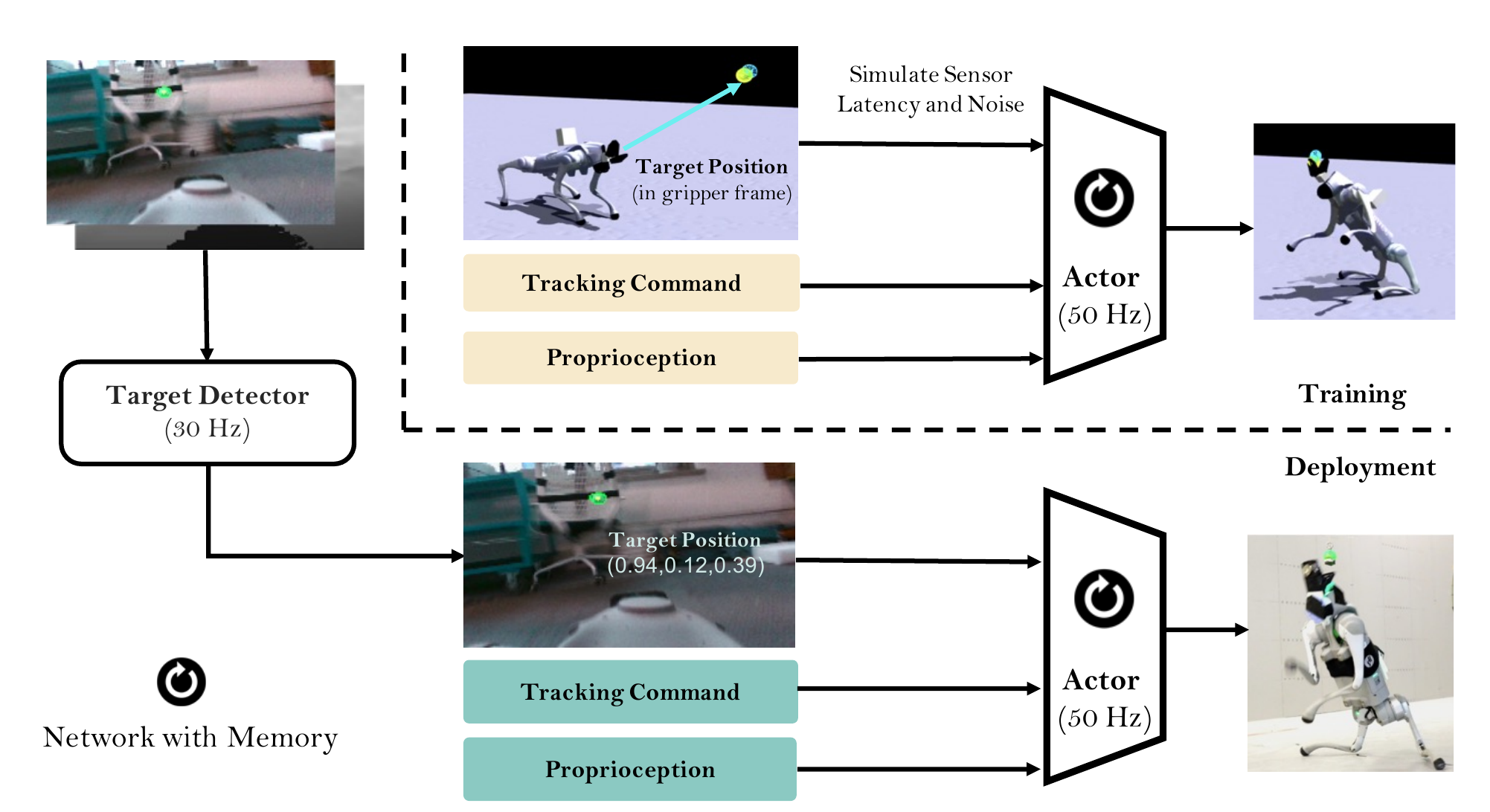}
\caption{\textbf{System Framework (Pipeline).} We use the policy network trained in the simulation to map the observation input, which includes proprioception and the goal position coordinates computed using the object detector, into goal joint angles. Then the PD controller computes the motor torques with respect to the goal joint angles, current joint angles, and joint velocities, and applies them to the real robot.}
\label{fig:pipeline}
\vspace{-0.4cm}
\end{figure*}
Our method trains a neural network that is mapping the proprioception, the compact target position and the desired robot velocity into joint angle commands. The general pipeline of our approach is shown in Figure~\ref{fig:pipeline}.

Decoupling the perception system allows each system to operate at its optimal frequency - e.g. our low-level controller runs at 50Hz while the perception pipeline can run at a lower frequency according to the hardware capability and the computation cost. This also allows us to model and handle perception uncertainty separately from control challenges during training.
Depth images are usually costly to render in the simulation and to process on board. Inspired by the training idea of using waypoints as the locomotion command~\cite{zhang2023learning}, we choose the relative position in the robot frame as the most compact representation of the target object, which provides adequate information to track the goal. We provide the desired velocity, as determined by the operator. The hope is, that the network learns to control the quadrupedal robot to track toward the object, following the given velocity value.

\subsection{Goal Oriented Reward}
During the RL training process, to motivate the robot to be agile and precise, we switch between two reward functions. We set the first tracking reward term $r_{\text {vel}}$ similarly to~\cite{cheng2023parkour}, based on the velocity command, encouraging the robot to follow the velocity command running toward the goal in an agile fashion. Additionally, we use the tracking yaw $r_{\text{tracking yaw}}$ reward term to help the robot face towards the goal.

In order for the robot to learn to also be precise, i.e. to accurately catch the target object, we set the $r_{\text {pos}}$ tracking reward based on the relative goal position in the robot end-effector frame, which exponentially increases with decreasing target-end-effector distance. We add a constant $1$ to make $r_{\text {pos}} \geq r_{\text {vel}}$, such that the robot wouldn't fall into some local optima while switching between the two reward functions.
Finally, a bit-wise reward indicates whether we successfully caught the object.

\begin{equation}
r_{\text{tracking goal}}=\left\{\begin{array}{ll}
    r_{\text {vel}}=\min \left(\left\langle\mathbf{v}, \hat{\mathbf{d}}_{w}\right\rangle, v_{c m d}\right), & d_{xy} > D \\
    r_{\text {pos}}=\exp{(-\left \|\mathbf{p}-\mathbf{x}  \right \|  / \alpha)}+1, & d_{xy} \leq D
\end{array}\right.
\end{equation}

\begin{equation}
    r_{\text{tracking yaw}}=\exp\left(-{|y_{goal} - y|}/{\sigma}\right)
\end{equation}

$\hat{\mathbf{d}}_{w}=\frac{\mathbf{p_{xy}}-\mathbf{x_{xy}}}{\|\mathbf{p_{xy}}-\mathbf{x_{xy}}\|}$ denotes the 2D goal direction vector in the xy-plane,
$x_{xy}$ is the robot base position and $p_{xy}$ is the goal position in the xy-plane in the world frame. $\mathbf{v}$ is the 2D base linear velocity in the horizontal plane and $v_{cmd}$ is the scalar velocity value we command. 
Regarding $r_{\text {pos}}$, $x$ is the 3D robot base position and $p$ is the goal position in the world frame.  $d_{xy}$ denotes the distance between the robot end effector and the target ball in the xy-plane, and $D$ is the threshold set to switch the tracking reward.
$y_{\text{goal}}$ is the target global yaw angle and $y$ is the current global yaw angle of the robot base.
$\alpha$ and $\sigma$ are scaling factors.

So $r_{\text {vel}}$ is rewarding the training to quickly move the robot towards the general goal direction, while $r_{\text {pos}}$ takes over when the robot is close to the goal object and strongly rewards high accuracy w.r.t. gripper-target distance by scaling exponentially, thus together training the network to be agile \textbf{and} precise.


We also use the regularization terms, including conserving mechanical energy as in~\cite{zhuang2023robot} to encourage reasonable biomechanically optimal gait. To overcome the gap of exploration, we set an easily-deployed curriculum for goal height. At the very beginning, the robot can easily capture the target balls almost on the same level as the base. As the robot get more successful (catching the ball and not falling over), our code will gradually choose to spawn more episodes with more difficult/ higher target positions, requiring the robot to learn to jump. However, it still takes some time for the robot to learn how to jump up and grab objects. This is because, at the beginning, the rewards obtained by the robot trying to jump up and grab the object were relatively sparse, so sometimes, the rewards obtained by stopping directly below the object reach a local optima. This can be solved by balancing the rewards for grabbing objects and distance rewards.

\subsection{Collision Shape and Grasping System in Simulation}
\begin{figure}[ht]
\vspace{-0.4cm}
\centering
\includegraphics[width=0.9\linewidth]{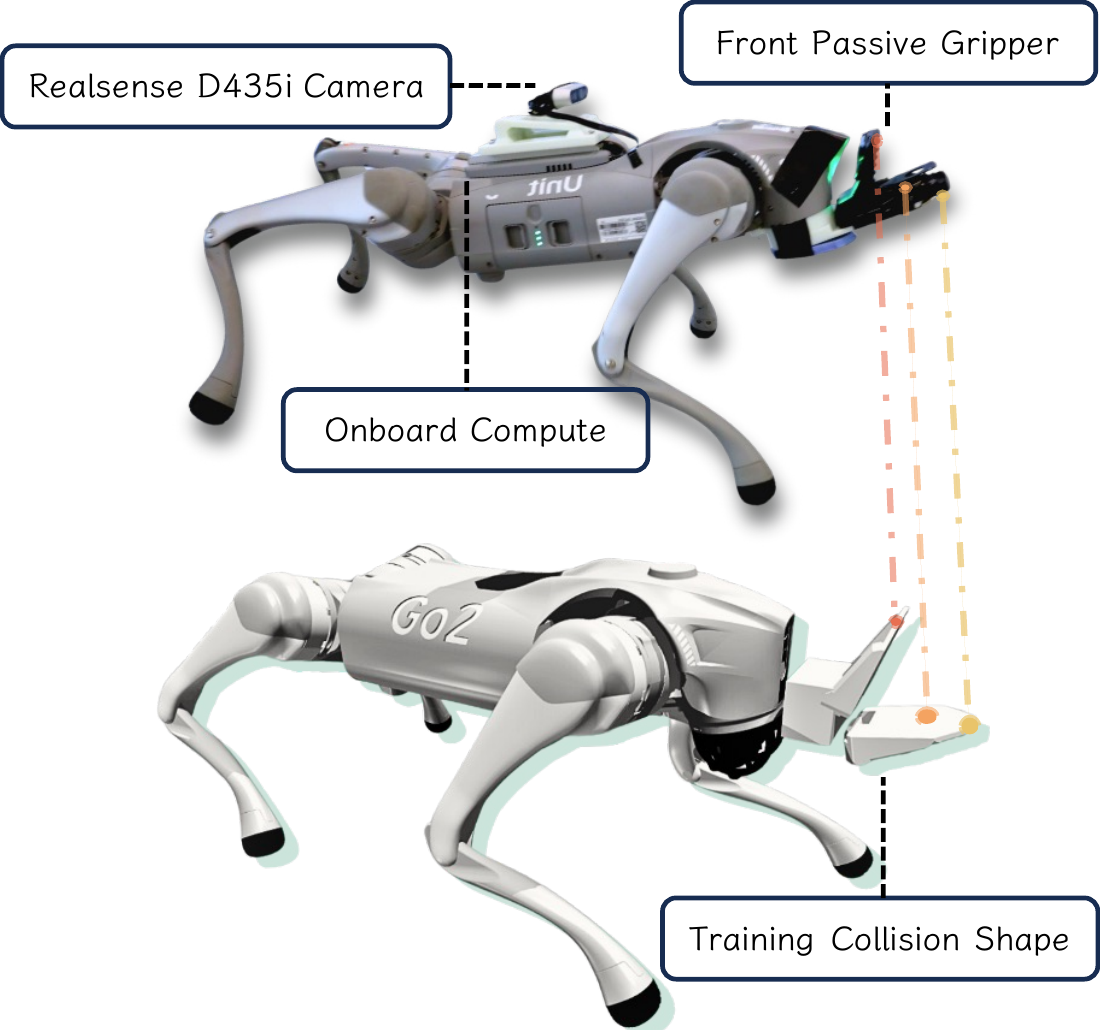}
\caption{\textbf{Hardware Setup.} Maintaining consistency in the collision shape of the front gripper between simulation and reality helps the robot learn to correctly trigger contact with the ball using the gripper at the appropriate position.\vspace{-0.1cm}}
\label{fig:gripper-collision}
\end{figure}
\paragraph{Collision Shape Setup}

The gripper on our actual robot is triggered via a passive mechanism to close when encountering a force induced by the ball hitting the gripper. For this reason, and to simplify the simulation, we do not simulate a gripper with an active joint that is cathing the ball by physically restraining it between the jaws. Instead, we use a simple catch PD controller to "catch" the ball in the mouth once it is very close to the center mouth point.



We apply mutual external forces between the ball and the gripper with a strong PD controller so that the gripper can hold the ball firmly once it reaches the correct grip, i.e. $\|p_\text{ball} - p_\text{gripper}\| < D_\text{grasp}$.

\begin{equation}
\begin{aligned} 
    F_\text{grasp} &= (K_\text{p, grasp}(p_\text{ball} - p_\text{gripper}) - K_\text{d, grasp}(v_\text{ball} - v_\text{gripper})) \\
    &* \mathbbm{1}[\|p_\text{ball} - p_\text{gripper}\| < D_\text{grasp}]
\end{aligned}
\label{eq:grasp-force}
\end{equation}
The mutual external force to the gripper is described as Equation~\ref{eq:grasp-force}, where $p_\text{ball}$ is the ball position in the world frame and $p_\text{gripper}$ is the middle point of the gripper in the world frame. The external force applied to the ball is the negative of $F_\text{grasp}$. To enforce a firm grasp, we set $K_\text{p, grasp} = 150.0$, $K_\text{d, grasp} = 2.0$.

\paragraph{Ball Flying Away vs. Not Flying Away}
Due to the high-speed dynamics of our task, in the simulation the ball tends to fly away and will never be caught once it collides with the gripper on the robot - even if it is colliding on the inside faces of the mouth. But we know from our real robot experiment, that the gripper will catch the ball even if it comes into the mouth with great velocity. Thus, we apply another correctional force to the ball with the catch PD controller to prevent the ball from flying away after being hit. 

Another problem is, that a collision of the ball with other parts of the robot model will caues it to fly far away. In order to keep the ball close, as if it would be hanging from a rope, we slowly move the ball back to its original position using a different PD controller. 
By tuning the stiffness and damping parameters of this controller, we can adjust the difficulty for the robot as well: In the beginning the ball comes back quickly, and later very slowly. In our experiments we have seen that this way the robot can learn its own collision information and better avoid collisions.



\subsection{System Pipeline}
After Reinforcement Learning using the rewards and methods outlined above, our system uses the trained policy network to convert observation inputs into goal joint angles, which the robot's PD motor controller then uses to calculate and apply motor torques to the real robot.

\paragraph{Target localization}
To detect the target, we utilize the RGB-D camera mounted on the back of the quadrupedal robot for exteroception, which captures images at 30 Hz. 
To simplify the training of the policy network and enhance its generalization, we represent the target information in the observation as the coordinate position of the object in the end-effector's frame. We first obtain the pixel coordinates of the target from the image, and then combine this with depth information and the camera's intrinsic parameters to derive the object's coordinates in the camera frame. Finally, using the camera's extrinsic parameters and its pose relative to the end-effector, we can calculate the target object's coordinate position in the end-effector frame. Additionally, there is a one-bit parameter that indicates whether the object has been detected. For training the simulator checks if the object is in the field of view of an assumed virtual camera to decide if the object pose is provided or not. 

Many off-the-shelf object detection methods could be used. However, we use the simplest HSV (Hue, Saturation, Value) detector in our real-world experiments, due to the real-time requirement and it working sufficiently well for our simple use-case. We extract the object's position within the image frame by utilizing openCV's ``findContours" function \cite{itseez2015opencv}. We then select the median depth value of all ranges in the biggest contour. Due to the high-speed motion of the robot, the collected images are prone to motion blur, resulting in the estimated contour not well overlapping with the pixels of the ball in the depth image, such that sometimes the calculated median depth is not on the ball but on the background. Therefore, we filter the estimated target distances of consecutive frames for big jumps in the distance.


\paragraph{Inference Network}
Considering that the dog may lose tracking with the ball while jumping and the latency in obtaining the target position, we employ a memory-equipped network, specifically a GRU (Gated Recurrent Unit), which processes the hidden state followed by a simple three-layer MLP (Multi-Layer Perceptron).

\paragraph{Deployment}
Our deployment on the real robot is based on ROS2 (Robot Operating System 2), which utilizes DDS (Data Distribution Service) for communication. ROS2 serves as a bridge connecting the policy network, robot node, and external sensors, enabling seamless integration of various modules and simplifying the collection of information from multiple sensors. 
The policy network converts the observation input, which includes proprioception and goal position coordinates obtained from the object detector, into goal joint angles at a frequency of 50 Hz. Subsequently, the PD motor controller calculates the motor torques based on the goal joint angles, current joint angles, and joint velocities, applying them to the real robot.
\begin{equation}
\tau = K_p * (q_{target}-q) - K_d * v
\end{equation}
where $q_{target}$ and $q$ are the target joint angles and current joint angles, $v$ denotes current joint velocities, and we tuned the parameters to $K_p=35$ and $K_d=0.6$. To ensure safe deployment, we implement a safety mechanism that terminates the program when the joint angles exceed the safe range or the angular velocity exceeds 55 degrees per second.
The HSV-based object detector runs at the camera frame rate of $30Hz$.

\section{EXPERIMENTS}
In order to validate our method and test the limits of our framework, we propose several experiments. We first investigate the necessity of the memory mechanism in our task using a trivial MLP, a GRU based network (our method), and a Transformer-based network. Then, we test how well our system estimates the target position by providing the truth value of the target height. By comparing these two setups, we evaluate the possible improvement of estimating the target position, as this is difficult while the robot is in motion (see below).


\subsection{Experiment Setup}
We deploy our policy on a Unitree Go2 robot with 12 joints and one extra mouth-like gripper with a total weight of 15kg. The height of the standing base is about 30cm and its body length is about 70cm. We run the policy on a Jetson Orin NX mounted on the robot.
For detecting the target, we use an Intel RealSense D435i mounted on the back of the quadrupedal robot for exteroception with a frame rate of 30Hz. Since our focus is not on hardware and our task requires the gripper to be able to instantly close and grasp objects, we design a passive gripper with strong magnets that can be swiftly triggered.

\subsection{Simulation Experiments}
\begin{figure*}
\centering
\includegraphics[width=0.9\linewidth]{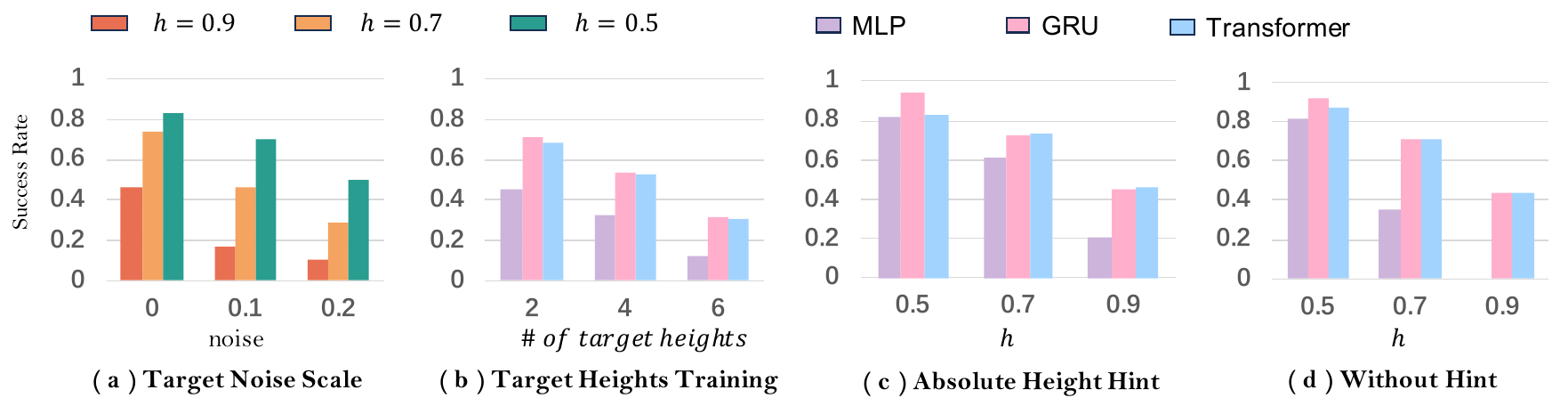}
\caption{\textbf{Comparisons.} (a) We examine the success rate of GRU for catching the target object while introducing uniform noise to the goal position at various heights. (b) We analyze the success rate of the training range for the number of different target heights during training across three different backbone architectures. (c) and (d) We evaluate the success rate based on whether the input includes the absolute target height, again comparing the performance of the three backbone models.}
\label{fig:compare}
\end{figure*}

\renewcommand{\arraystretch}{1.1}
\begin{table}[htbp]
\centering
\begin{tabular}{|c|c|c|c|c|c|c|}
    \hline
    \textbf{} & \multicolumn{6}{|c|}{\textbf{Success Rate}} \\
    \cline{2-7}
    \textbf{Policy} & \multicolumn{3}{|c|}{\text{$h$ (Simulation)}} & \multicolumn{3}{|c|}{\text{$h$ (Real-World)}} \\
    \cline{2-7}
    \textbf{Backbone} & 
    \textbf{0.5m} & 
    \textbf{0.7m} &
    \textbf{0.9m} & 
    \textbf{0.5m} & 
    \textbf{0.7m} &
    \textbf{0.8m} \\
    \hline
    \textbf{MLP} & 
    \text{0.819} & 
    \text{0.354} &
    \text{0.000} & 
    \text{-} & 
    \text{-} & 
    \text{-} \\
    \hline
    \textbf{GRU} & 
    \text{0.922} & 
    \text{0.708} &
    \text{0.433} & 
    \text{9/10} & 
    \text{3/10} &
    \text{1/10} \\
    \hline
    \textbf{Transformer} & 
    \text{0.870} & 
    \text{0.711} &
    \text{0.437} & 
    \text{8/10} & 
    \text{3/10} &
    \text{2/10} \\
    \hline
\end{tabular}
\caption{\textbf{Success Rate Experiments.} We quantitatively test the success rates of three policy networks with MLP, GRU, and Transformer as backbones for target heights of 0.5m, 0.7m, and 0.9m/ 0.8m in simulation/ real-world environments.}
\label{tab:exp_real}
\end{table}

\renewcommand{\arraystretch}{1.}

\subsubsection{Success rate}
A robots's attempt is classified as success if it jumps, catches the target ball, and then lands without falling over. We consider it successful if the robot catches the ball within three attempts of its front legs leaving the ground, as the robot sometimes requires real-time adjustment information to better know the desired jumping height.
We test the success rate of the robot at different target heights from 0.5m to 0.9m in the simulation environment, as shown in Table \ref{tab:exp_real}. We also present keyframe animations of the robot grasping different target objects in Figure \ref{fig:main image}.
Due to the significant impact of the allowable torque on the robot's jumping ability in the simulation, failing to impose a torque limit could result in a large sim-to-real gap, making the success rates in the simulation less meaningful. Thus, we set a maximum torque limit of 30 Nm for the robot in the simulation. We selected the checkpoints of the three best-performing corresponding policy networks for testing. 
The policy networks based on MLP, GRU, and Transformer achieve success rates of over 80\% at a target height of 0.5m, with MLP performing the worst. 
However, when the target height increases to 0.7m, the success rate decreases as the task difficulty rises, with MLP dropping to around 35\%, while GRU and Transformer maintained success rates above 70\%. As the target height further increases, the MLP struggles due to its lack of memory, resulting in failure at 0.9m, while both the GRU and Transformer achieve success rates exceeding 40\%. The overall performance of the GRU and Transformer is quite similar. During the training process, we found that as the required grabbing height increases from a level that did not require jumping to a height that did, a lack of careful adjustment of the reward could result in the model failing to learn that it could jump, causing it to get stuck in a local optimum. The policy based on the Transformer architecture demonstrates some advantages in escaping from this local optimum.

\begin{figure}[ht]
\centering
\includegraphics[width=\linewidth]{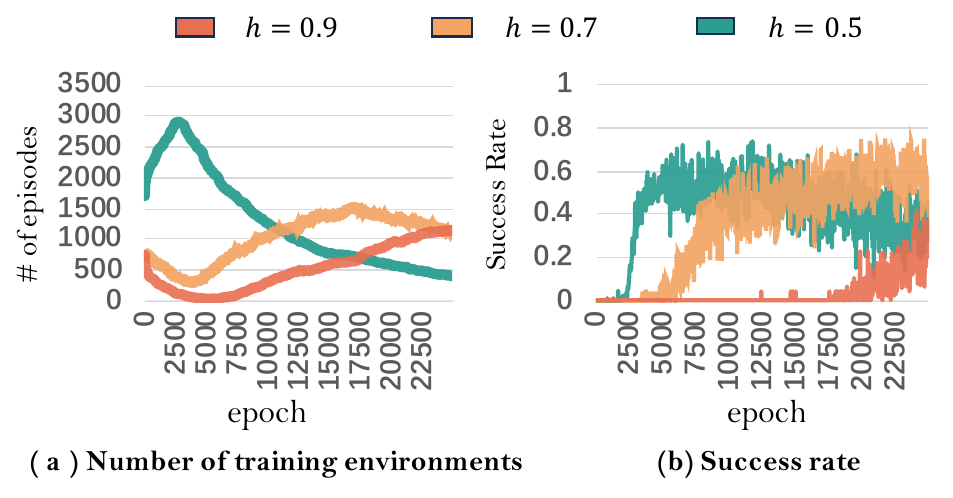}
\caption{\textbf{Curriculum Learning.} When broadening the height range of objects that robots can grasp through curriculum settings, we observe that the success rate for a specific height may decline due to a reduced number of episodes associated with that height.}
\label{fig:multimodality}
\end{figure}

\subsubsection{Curriculum Learning}

During the training process, we use a curriculum to adjust the height of the target objects, preventing the agent from giving up on catching the ball when faced with difficult targets.
When expanding the height range of objects grasped by robots by curriculum setting, we noticed that the success rate of a specific height may decrease due to the decrease in the number of environments corresponding to this height. In one training session shown in Figure~\ref{fig:multimodality}, we set up 4096 parallel environments and applied height curriculum to dynamically adjust the task difficulty. During the training process, while the number of environments set to a height of 0.5m decreases to 25\% of the total number of environments, the success rate at that height exhibits a downward trend. 

As shown in Figure~\ref{fig:compare} (b), as the number of different heights that the network is trained to master increases, the average success rate goes down. 
%
%
We then hypothesize that this may  result from the robot's potential pitch during movement, which makes the representation of the target's height using relative position less clear. Therefore, we introduce a hint of the target object's absolute height into the robot's observation input to compare the impact of this hint on the success rate.
Figures~\ref{fig:compare} (c) and (d) illustrate the two scenarios with and without the absolute height hint. However, the inclusion of the hint significantly impacts the success rate only for the MLP backbone at heights above 0.7m, which may be attributed to MLP's lack of memory capability. We speculate that this may also be related to the different gaits the robot exhibits when grasping objects at varying heights, as shown in Figure~\ref{fig:main image}. After the curriculum updates, training on a new target may have affected the model's performance on the original target. This is also a point we can explore further in our future work.

\subsection{Real-World Experiments}
We also quantitatively test our policy in the real world. 
We select the best-performing checkpoints based on GRU and Transformer backbones from the simulator for testing on the Unitree Go2 real robot. The success rates of different heights of the target ball over 10 trials each are shown in Table~\ref{tab:exp_real}. 
When the target object's height is 0.5m - just slightly above the robot's standing height - the success rate of grasping the object is very high, more than 80\%. However, as the height of the object increases, the difficulty of the task significantly rises, leading to a corresponding decrease in the success rate. 

Due to the significant noise and latency in the relative position information of the target object captured by the camera while the robot is in high-speed motion, the success rate in the real world is lower compared to experiments at the same height conducted in the simulation.
When the target height is 0.7m, the robot's success rate is approximately 30\%; however, when the target height increases to 0.8m, the success rate drops to 10\%. 

There exists a sim-to-real gap, as the real robot cannot achieve the highest height successfully reached in the simulation.
The real robot can catch the highest ball at 0.8m, roughly twice the height at which the robot is standing. During the experiments, we found that minimizing the difference in camera latency between the real world and training, by adding a delay to providing the simulated object pose, does improve the success rate of target grabbing. The model that performs best on the real robot has a latency sample range of [0.03, 0.08] seconds during the training process. We test our policy in several different indoor and outdoor environments, including relatively smooth surfaces, grassy areas, and uneven outdoor stone pathways. The robot can catch a fixed ball or play with a moving ball held by a human, demonstrating the ability to continuously follow a mid-air ball moving at speeds of up to 3m/s with direction changes, maintaining a close pursuit without losing track of the target, as shown in Figure~\ref{fig:main image} and the video.



\section{CONCLUSION, LIMITATION and FUTURE WORK}
In this work, we demonstrate the capability of a quadruped robot equipped with an end-effector designed to mimic a dog's mouth to jump and catch small objects. By training the robot in simulation, we achieve notable performance, allowing it to capture targets at various heights in the simulation, and with the highest height of 0.8m real-world deployment. 
Based on the experiments, we show that the robot exhibits different gaits when grasping objects at different height levels.
Additionally, there is a noticeable gap between the results in real-world scenarios and those in simulation. This discrepancy may stem from hardware limitations, the sim-to-real gap, and noise from the existing detector during high-speed motion. These factors represent key areas for future research aimed at enhancing the robot's performance in high-dynamic movements.

We also did some preliminary work to enable the robot pursue and catch arbitrary objects using the Grounding DINO~\cite{liu2023grounding} open-vocabulary detector and Mobile SAM~\cite{mobile_sam} tracker. This pipeline works on the Nvidia Jetson NX at $8\pm 2$ Hz. Although this frequency is not very sufficient for the robot to accurately catch when needing to jump up, it shows our ability to track different objects and play with humans. 
This indicates that our agile locomotion skills can be integrated with the system based on vision-language models (VLMs), such as in a helping task~\cite{wu2024helpful}, allowing us to perform a broader range of tasks, which we will pursue in the future.



\section*{ACKNOWLEDGMENT}

This project is supported by Shanghai Qi Zhi Institute and ONR grant N00014-20-1-2675 and has been partially funded by the Shanghai Frontiers Science Center of Human-centered Artificial Intelligence. The experiments of this work were supported by the core facility Platform of Computer Science and Communication, SIST, ShanghaiTech University. Thanks to Qi Wu for the detailed technical support. We thank Zipeng Fu, Xuxin Cheng, Wenhao Yu and Erwin Coumans for the feedback and discussion.


 \IEEEtriggeratref{39}
\bibliographystyle{unsrt}
\bibliography{main}

\end{document}